\documentclass[conference]{IEEEtran}
\IEEEoverridecommandlockouts
\usepackage{cite}
\usepackage{amsmath,amssymb,amsfonts}
\usepackage{graphicx}
\usepackage{textcomp}
\usepackage{xcolor}
\def\BibTeX{{\rm B\kern-.05em{\sc i\kern-.025em b}\kern-.08em
    T\kern-.1667em\lower.7ex\hbox{E}\kern-.125emX}}

\usepackage{supertabular}

\usepackage[utf8]{inputenc}
\usepackage{listings}
\usepackage{color}
\usepackage{algorithm}
\usepackage{algpseudocode}
\usepackage{subcaption}
\usepackage{multicol}

\definecolor{dkgreen}{rgb}{0,0.6,0}
\definecolor{gray}{rgb}{0.5,0.5,0.5}
\definecolor{mauve}{rgb}{0.58,0,0.82}

\renewcommand\figurename{Matrix}

\begin{document}

\title{Predicting the Score of Atomic Candidate OWL Class Axioms} 

\author{\IEEEauthorblockN{Ali Ballout}
\IEEEauthorblockA{\textit{Universit{\'e} C{\^o}te d’Azur, I3S, Inria} \\
Sophia Antipolis, France \\
ali.ballout@inria.fr}
\and
\IEEEauthorblockN{Andrea G. B. Tettamanzi}
\IEEEauthorblockA{\textit{Universit{\'e} C{\^o}te d’Azur, I3S, Inria} \\
Sophia Antipolis, France \\
andrea.tettamanzi@unice.fr}
\and
\IEEEauthorblockN{C{\'e}lia da Costa Pereira}
\IEEEauthorblockA{\textit{Universit{\'e} C{\^o}te d’Azur, I3S, CNRS} \\
Sophia Antipolis, France \\
celia.pereira@unice.fr}}

  

\maketitle

\begin{abstract}
Candidate axiom scoring is the task of assessing the acceptability of a candidate axiom against the evidence provided by known facts or data.
The ability to score candidate axioms reliably is required for automated schema or ontology induction, but it can also be valuable for ontology and/or knowledge graph validation.
Accurate axiom scoring heuristics are often computationally expensive, which is an issue if you wish to use them in iterative search techniques like level-wise generate-and-test or evolutionary algorithms, which require scoring a large number of candidate axioms.
We address the problem of developing a predictive model as a substitute for reasoning that predicts the possibility score of candidate class axioms and is quick enough to be employed in such situations.
We use a semantic similarity measure taken from an ontology's subsumption structure for this purpose.
We show that the approach provided in this work can accurately learn the possibility scores of candidate OWL class axioms and that it can do so for a variety of OWL class axioms.
\end{abstract}





\section{Introduction and Related Work}
\label{introduction}

An ontology is the explicit representation of components of a shared conceptualization~\cite{GruberR.1995}. In machines, an ontology is a vocabulary which is used by said machine in the representation of knowledge. An AI knowledge-based system would take an ontology as its universe of components and their relations and derive new implicit knowledge within that universe.

As for the components of an ontology,  Z. Dragisic~\cite{Dragisic2017} defines them as follows:
{\small
\begin{itemize}
  \item Concepts (classes) - the types of objects in a domain or area.
  \item Relations (properties, roles) - the relations between two concepts/classes.
  \item Instances (individuals) - instances of concepts/classes.
  \item Axioms - model sentences that are always correct in the domain. They're often utilized to represent information that cannot be formally specified by the other components~\cite{Corcho2006}.
\end{itemize}
}
Ontologies play a vital role in data and knowledge integration by making a common schema available~\cite{Hadzic2009}. Unfortunately, ontology construction is extremely expensive, in terms of both time and resources, and dependent on the availability of knowledge experts~\cite{Hadzic2009,Simperl2012}.
The construction of an ontology for a certain field requires the contribution of a knowledge engineer and of an expert in that specific field. This dependency persists throughout the lifetime of an ontology as an expert is required to develop and expand it as new requirements arise.
To overcome such a bottleneck in knowledge acquisition, the field of ontology learning was conceived.
Ontology Learning~\cite{Hadzic2009,LehmannVoelker2014,MaedcheStaab2001} is the task consisting of the automatic generation of ontologies. Ontology learning includes a variety of techniques and those are grouped into:
{\small
\begin{itemize}

  \item Linguistic techniques - Natural language processing mostly used in the  preprocessing of data, and some learning tasks such as term extraction~\cite{Asim2018}.
  \item Statistical techniques - such as data mining and information retrieval methods used to extract terms and associations between them~\cite{FleischhackerVoelker2011,FleischhackerVoelkerStuckenschmidt2012}.
  \item Inductive logic programming (ILP) is a branch of machine learning that uses logic programming to generate hypotheses based on prior knowledge and a set of examples~\cite{FanizziDAmatoEsposito2008,Lehmann2009,ILPat20}. 

\end{itemize}
}
Each one of these technique groups is involved in one or more of the stages of ontology learning, those stages being preprocessing, term and concept extraction, relation extraction, concepts and relations hierarchies, axioms schemata and general axioms~\cite{Asim2018}. Linguistic techniques can have a role in almost all the stages, but as we mentioned they are mostly used for data preprocessing. 

As for statistical techniques, they are methods that rely entirely on the statistics of the textual resources without any concern for the semantics. One such method is the possibilistic (statistics-based) heuristic detailed in~\cite{Tettamanzi2017}, where the authors have developed a possibilistic framework for the Web ontology language (OWL 2) to test axioms against evidence expressed in the Resource Description Framework (RDF). The heuristic uses support, confirmations, and counterexamples to define possibility and necessity of an axiom and an acceptance/rejection index combining both of them. They test the developed theory on \verb!SubClassOf! axiom testing against the DBpedia\footnote{https://www.dbpedia.org/resources/ontology/} database. The results of their experiments showed that the method was suitable for axiom induction and ontology learning~%
\cite{Tettamanzi2014}. The suggested heuristic has the drawback of being
time consuming; 
this was addressed in a revision of the method that added a time cap for the querying process, of course at the cost of a little increase in error rate reaching 3.96\%~\cite{Tettamanzi2015}. The number of axioms tested was 5050 and it took almost 342 CPU hours and a half with an average of 244 s per axiom \textbf{with time capping}. This is a significant improvement, considering that testing the same amount of axioms without a time cap would have taken approximately 2,027 CPU days. Another limitation is that lack of support means an inconclusive judgement, as the method queries for confirmations and counterexamples, and sometimes it might find none. The third drawback is that being a statistical approach it relies solely on instance data from the dataset, data that is prone to errors. 

In comparison, ILP techniques are a sub-field of machine learning that follow exhaustive statistical or linguistic processing. One such example is~\cite{Lisi2013}, where the authors describe a method for the automated induction of fuzzy ontology axioms which follows the machine learning approach of ILP named \verb!SoftFOIL!. One of \verb!SoftFOIL!'s limits is a result of its sequential covering strategy. As it uses a greedy search to find rules, it does not guarantee to find the smallest or best set of rules that explain the training examples. Another is susceptibility to being trapped in a loop while searching for the best rule.

A new emerging group of techniques, is a hybrid breed that takes advantage of combining classical ILP and statistical machine learning. To stay in the context of the previous examples, a model would be trained with axioms having their scores assigned by a statistical method such as~\cite{Tettamanzi2017}, as well as using a similarity measure that results from the logical processing of the data. This is exactly what was done in~\cite{Malchiodi2018}, where the authors modified the support vector clustering algorithm to attempt to predict the possibilistic score of \textbf{OWL} axioms. They used the heuristic from~\cite{Tettamanzi2017} as the scorer, and used a model originally developed for inferring the membership function for fuzzy sets. As a similarity measure they used one specific for subsumption axioms based on semantic considerations and reminiscent of the Jaccard index. The predictor performance was poor in terms of root mean square error (RMSE) scoring 0.572 (Table~3 in~\cite{Malchiodi2018}). In addition, the authors mentioned that they found a group of axioms that were \emph{hard} to predict, and could not find a reason that explains why. This method was computationally far more efficient than~\cite{Tettamanzi2014}, yet it was still reliant on the instances in the data set and querying them to construct the similarity measure, meaning that even though it is an improvement, it still falls victim to the same problems. The authors explicitly mention that a major weakness of their method is that training such a model consumes a significant amount of resources.

Our work addresses the shortcomings of the previous techniques that heavily rely on error-prone instance-dependent statistics. It is also able to predict the scores of multiple types of axioms and is not bound to simply \verb!subsumption!. We propose a method that can be used as a building-block or an extension/plug-in to other existing statistical analysis or ILP options, such as DL-Learner~\cite{BUHMANN201615}, to allow faster execution while maintaining high scoring accuracy, while still having the ability to perform as a simpler stand-alone scorer. The method works by training a model on a set of atomic class axioms scored by an algorithm, in this case~\cite{Tettamanzi2017}. This enables the model to predict the score of any new atomic (consisting of a single concept on each side) candidate axiom. We experimented using multiple machine learning methods, and compared our work to the state of the art~\cite{Malchiodi2018} that aims to achieve the same goal.

This paper is structured as follows: Sect.~\ref{background} provides some background about both axiom scoring and concept semantic similarity which are both prerequisites to training the models. As for Sect. ~\ref{methodology} it lays out the methodology explaining how the axioms were extracted and scored, how the semantic measure we use was developed, and also how an axiom based vector space was modeled leading to the prediction of the scores. We detail our experiments including a comparison with the method presented in~\cite{Malchiodi2018} in Sect.~\ref{experiments} then present the results while listing our observations and findings. We end the paper with some notes and conclusions.

\section{Background}
\label{background}

\subsection{Axiom Scoring}
\label{score}
Axiom scoring can be done using statistical analysis, we mentioned in the previous section a heuristic~\cite{Tettamanzi2017} that does this using the theory of possibility. Possibility theory~\cite{DuboisPrade1988} is a mathematical theory of epistemic uncertainty.
Its central notion is that of a possibility distribution which assigns to each elementary event a degree of possibility ranging from 0 (impossible, excluded) to 1 (completely possible, normal). A possibility distribution $\pi$ induces a \emph{possibility measure} $\Pi$, corresponding to the greatest of the possibilities associated to an event and the dual \emph{necessity measure} $N$, equivalent to the impossibility of the negation of an event.


Here we provide a brief explanation of the theory behind the scoring. Given a candidate OWL~2 axiom $\phi$, expressing a \emph{hypothesis} about the relations
holding among some entities of a domain, we wish to evaluate its credibility,
in terms of possibility and necessity, based on the \emph{evidence} available
in the form of a set of facts contained in an RDF dataset $\mathcal{K}$.

The content of $\phi$ is defined as a (finite) set of \emph{basic statements}
$\psi$ which are logical consequences of $\phi$, i.e., $\phi\models\psi$.
The open-world hypothesis (OWA) is fully taken into account. Therefore,
given $\mathcal{K}$, for each such $\psi$, there are three cases:
\begin{enumerate}
\item $\mathcal{K} \models \psi$:
  in this case, we will call $\psi$ a \emph{confirmation} of $\phi$;
\item $\mathcal{K} \models \neg\psi$:
  if so, we will call $\psi$ a \emph{counterexample} of $\phi$;
\item $\mathcal{K} \not\models \psi$ and $\mathcal{K} \not\models \neg\psi$:
  in this case, $\psi$ is neither a confirmation nor a counterexample of $\phi$.
\end{enumerate}

If we denote by $u_\phi$ the \emph{support} of $\phi$, which is the cardinality
of its content, by $u_\phi^+$ the number of confirmations of $\phi$
and by $u_\phi^-$ the number counterexamples of $\phi$,
the possibility and the necessity of candidate axiom $\phi$ may be defined as follows:
\begin{itemize}
  \item if $u_\phi > 0$, 
{\small
\begin{eqnarray}
  \label{eq:poss-conj}
  \Pi(\phi) &=& 1 - \sqrt{1 - \left(\frac{u_\phi - u_\phi^-}{u_\phi}\right)^2}; \\
  \label{eq:nec-conj}
  N(\phi) &=& \left\{\begin{array}{ll}
    \sqrt{1 - \left(\frac{u_\phi - u_\phi^+}{u_\phi}\right)^2},\quad & \mbox{if $u_\phi^- = 0$,}\\[1.5em]
    0, & \mbox{if $u_\phi^- > 0$;}
  \end{array}\right.
\end{eqnarray}
}
\item if $u_\phi = 0$, $\Pi(\phi) = 1$ and $N(\phi) = 0$,
  i.e., we are in a state of maximum ignorance, given that no evidence is available
  in the RDF dataset to assess the credibility of $\phi$.
\end{itemize}

The possibility and necessity of an axiom can then be combined into
a single handy acceptance/rejection index
\begin{equation}\label{eq:ARI}
{\small
  \begin{array}{rcl}
    \mathrm{ARI}(\phi) &=& N(\phi) +\Pi(\phi) -1 = N(\phi) - N(\neg\phi)\\
       &=& \Pi(\phi) - \Pi(\neg\phi) \in [-1, 1],
  \end{array}}
\end{equation}
because $N(\phi) = 1 - \Pi(\neg\phi)$ and $\Pi(\phi) = 1 - N(\neg\phi)$
(duality of possibility and necessity).
A negative $\mathrm{ARI}(\phi)$ suggests rejection of $\phi$ ($\Pi(\phi)<1$),
whilst a positive $\mathrm{ARI}(\phi)$ suggests its acceptance ($N(\phi)>0$),
with a strength proportional to its absolute value. A value close to zero
reflects ignorance about the status of $\phi$.

\subsection{Ontological Semantic Similarity}
\label{similarity}
Semantic similarity is a notion used to define a distance between terms or concepts based on meaning or semantics. It includes in its calculation only relations of the type~\textbf{IS-A}~\cite{Ballatore2014}. It is often confused with semantic relatedness, for example a~\textit{train} and~\textit{train tracks} are functionally complementary, where as a~\textit{train} and an ~\textit{airplane} are functionally similar. The latter is an instance of semantic similarity where the relatedness of both terms is based on the defining features they share where both are vehicles~\cite{Ballatore2014,Corby2006}. Most semantic similarity measures that rely on a structured ontology, are based on path lengths between concepts as well as depth of concept nodes in an~\textbf{IS-A} hierarchy. As for information-based measures they use information content (\textbf{IC}) of concept nodes derived from the ontology hierarchy structure and corpus statistics~\cite{Al-Mubaid2006}.

The similarity measure we utilize in our work is the one detailed in~\cite{Corby2006-1,Corby2006}, under the subsection titled~\emph{Ontological Approximation}. The idea is to calculate the ontological distance between two concepts by using the subsumption path length. Following is the general definition:


{\small
\begin{equation}\label{eq:ontdis}
\begin{gathered}
    \forall(t_{1},t_{2})\in H^{2},\\ 
    D_{H}(t_{1},t_{2}) = min_{t}(l_{H}(\left<t_{1},t\right>)+l_{H}(\left<t_{2},t\right>))\\ 
     = min_{t}\left(\sum_{\{x\in<t_{1},t>,x\neq t_{1}\}}1/2^{d_{H}(x)}  +  \sum_{\{x\in<t_{2},t>,x\neq t_{2}\}}1/2^{d_{H}(x)}\right)
\end{gathered}
\end{equation}}
    
Formula~\ref{eq:ontdis} translates to: for all type pairs $t_{1}$ and $t_{2}$ in an inheritance hierarchy \emph{H}, the \emph{ontological distance} between $t_{1}$ and $t_{2}$ in the inheritance hierarchy \emph{H} is the minimum of the sum of the lengths of the subsumption paths between each of them and a common super type. And the length of the subsumption path between a type $t_{1}$ and its direct super type $t$ is equal to $1/2^{d_{H}(t)}$ with $d_{H}(t)$ being the depth of t in \emph{H}.

The authors of ~\cite{Corby2006} implemented this measure as part of a larger ontology-based search engine tool named \verb!Corese!~\cite{CorbyDF2004}\footnote{https://project.inria.fr/corese/}. This is the tool that has been used as means of extracting the semantic similarity between concepts in our method.

We propose an extension to this similarity to present similarities between axioms, this process is detailed in Sect.~\ref{axiomsimilarity}.

\subsection{Instance Semantic Similarity}\label{sec:axiom-similarity}

This similarity is used in the state of the art method detailed in~\cite{Malchiodi2018}. The support vector clustering method the authors used requires a kernel function which was assumed to return the similarity between two axioms.

Similar to the ontological similarity, it is based on the semantics of axioms and not on their syntax. The measure $S$ is defined with values in $[0, 1]$,
 satisfying the following desirable properties:
 for all axioms $\phi$ and $\psi$,
 {\small
 \begin{enumerate}
 \item $0 \leq S(\phi, \psi) \leq 1$;
 \item $S(\phi, \psi) = 1$ if and only if $\phi \equiv \psi$;
 \item $S(\phi, \psi) = S(\psi, \phi)$.
 \end{enumerate}}
 
 The similarity is defined based on a fuzzy implication operator $\implies$: we can say two axioms $\phi$ and $\psi$ 
are similar to the extent that $\phi \Rightarrow \psi$ and $\psi \Rightarrow \phi$.

 
 A fuzzy definition, based on the Herbrand semantics of the axioms,
 might be the following:
  {\small
 \begin{equation}\label{eq:fuzzy-implication}
  \implies(\phi, \psi) = \frac{\|\{\mathcal{I} : \mathcal{I} \models
 \neg\phi \lor \psi\}\|}{\|\Omega\|} = \frac{\|[\neg\phi] \cup
 [\psi]\|}{\|\Omega\|} = \frac{\|\overline{[\phi]} \cup
 [\psi]\|}{\|\Omega\|},
 \end{equation}}
 where $[\phi]$ denotes the set of the models of $\phi$.\footnote{A model of $\phi$ is an interpretation that satisfies $\phi$ and a counter-model is an interpretation that does not satisfy $\phi$.}

 One problem is that instead of exactly
 computing the numerator, which would require to count the models and counter models of both
 axioms being compared, a not so convincing rough approximation of it was used. This approximation was obtained by replacing models and counter models with instances occurring in the RDF dataset that confirm or contradict
 the two axioms. This renders the similarity bound to and limited by instance data, which means it is prone to errors as well as unusable in case instance data is scarce or non existent.

The similarity addresses and works with subsumption axioms only, of the form $C \sqsubseteq D$,
 where $C$ and $D$ are two OWL class expressions, and their negation $C \not\sqsubseteq D$.
 Given an individual $a$ occurring in an RDF dataset, we may say that
  {\small
 \begin{itemize}
 \item $a$ confirms axiom $C \sqsubseteq D$ (contradicts $C
 \not\sqsubseteq D$)
   iff $C(a) \land D(a)$;
 \item $a$ contradicts axiom $C \sqsubseteq D$ (confirms $C
 \not\sqsubseteq D$)
   iff $C(a) \land \neg D(a)$.
 \end{itemize}}
 Instead of counting the models of an axiom, the
 individuals
 that confirm it are counted; instead of counting its counter models, the individuals
 that contradict it.
 Given four OWL class expressions $A$, $B$, $C$, and $D$, the similarity which is also the degree
 to which axiom $A \sqsubseteq B$ implies axiom $C \sqsubseteq D$ can be computed as
 {\small
 \begin{equation}
 \label{eq:denom}
   S(A \sqsubseteq B, C \sqsubseteq D) =
   \frac{\| [A] \cap [B] \cup [C] \cap [D]\|}{\|[A] \cup [C]\|}.
 \end{equation}}%
In order to predict the ARI of subsumption axioms in this case, similarities between positive and negated subsumption axioms need to be computed.

 The similarity between two candidate OWL axioms of the form
 $A \sqsubseteq B$ and $C \sqsubseteq D$,
 can be computed using SPARQL counting queries.
 For instance, the denominator $\|[A] \cup [C]\|$ may be computed by
 {\small
 \begin{equation}\label{eq:querysim}
   \begin{minipage}[c]{5in}
     \begin{tabbing}
       \quad\=\quad\=\quad\=\kill
       \texttt{SELECT (count(DISTINCT ?x) AS ?n)}\\
       \texttt{WHERE} \{ \{ \texttt{?x a} $A$ . \} \texttt{UNION}
                         \{ \texttt{?x a} $C$ . \} \},
     \end{tabbing}
   \end{minipage}
 \end{equation}}%
whereas the numerators (which are four when comparing two axioms and their negations)
may be computed by SPARQL queries of the form

 \begin{equation}\label{eq:querysim2}
   \begin{minipage}[c]{5in}
{\small
     \begin{tabbing}
      
       \texttt{SELECT (count(DISTINCT ?x) AS ?n)}\\
       \texttt{WHERE} \{ \=\{ $Q([A])$ . $Q([B])$ . \}\\
       \texttt{UNION}\{
       \{ $Q([C])$ . $Q([D])$ . \} \},
       
     \end{tabbing}
}
   \end{minipage}
 \end{equation}
 where
  {\small
 \begin{eqnarray*}
   Q([X]) &=& \mbox{\texttt{?x a} $X$}, \\
   Q(\overline{[X]}) &=& \mbox{\texttt{FILTER NOT EXISTS ?x a} $X$}.
 \end{eqnarray*}}

This is a slow process, and for every candidate axiom you would need to calculate the similarity for it and its negation. Since the queries used are counting queries similar to those used in the scoring heuristic this means it also suffers from the same limitations such as heavy computation cost and instance dependency.


\section{Methodology}
\label{methodology}
In an~\textbf{OWL} ontology containing an inheritance hierarchy of concepts formed by the subsumption axiom \emph{rdfs:subClassOf}, our aim is to predict an acceptability score for a candidate atomic class axiom by learning a set of previously scored axioms of the same type, the score used is the one detailed in~\ref{score}. A model is built for each type of axiom to be predicted, this means that the method is repeated for the number of axiom types being dealt with. To measure the similarity between (candidate) axioms, we construct a similarity measure by extending the \emph{ontological distance} discussed in~\ref{similarity}, which is defined among concepts, not axioms. To this end, we consider the following necessary steps:

\begin{enumerate}
    \item \textbf{Axiom extraction and scoring}: This step constitutes the creation of the set of scored axioms of a certain type to be learned. We either use a ready scored set such as we did for our comparison, or we score a new generated set.
    \item \textbf{Semantic measure construction and assignment}: This step involves the retrieval of concepts used in our set of axioms, and their ontological distance from the ontology. Followed by extending that similarity to those axioms. This was done by calculating a single value that represents the similarity between each pair of axioms, by applying a function such as~\emph{Average} to the ontological distances of concepts in those axioms.
    \item \textbf{Axiom base vector space modeling}: This step focuses on using axiom similarity measures as weights, each axiom can be represented as a vector in an axiom based vector space. 
    \item \textbf{Score prediction}: This step is dedicated to training a Machine Learning model with the data set (vector space model in  addition to the scores) and predicting the acceptability score of new candidate axioms.
\end{enumerate}

We start by collecting the set of scored axioms we plan to train and test our method with. After that, we query the ontology to retrieve the ontological distances between all concepts. Then similarity measures between axioms will be derived from those distances and assigned as weights to represent the axioms as vectors in an axiom-based vector space. Machine learning models are used in the end to learn the set of scored axioms with their similarity weights and predict the acceptability score of a candidate axiom. 
We will consider subClassOf axioms for the comparison with~\cite{Malchiodi2018} since that is the only type of axiom they can address, and their dataset which we use for said comparison is made of subClassOf axioms. We will also use disjointWith axioms for our experiment to show that we can apply our method to all atomic owl class axiom types, as well as highlight that no leakage or bias is present from utilizing the subclass of hierarchy.

\subsection{Axiom Extraction and Scoring}
\label{axiomextraction}
In this paper, we consider two approaches. First generating a number of atomic class candidate axioms randomly. The generated candidates are filtered for duplicates. We then make sure non of the candidates already exist in the ontology explicitly or implicitly, using the search engine and its reasoner, then we score them.  

The other, which we prefer and use for controlled tests is to query existing axioms. To make sure we have positive scores we query the ontology for existing axioms and score them. The same can be said for negatively scored axioms, we can query the counter type and score it as if it were the first. For example subClassOf and disjointWith are counter types so if $disjointWith(C_{1} C_{2})$ has a positive score, $subClassOf(C_{1} C_{2})$ will have a negative one. This is bound by how many axioms exist to be queried (after inferring implicit axioms).

Query~\ref{lst:axiomextraction} is used to extract existing axioms of both types and labeling them to be scored after with the heuristic. The search engine used is Corese~\cite{CorbyDF2004} and the ontology is Dbpedia. After the extraction of the axioms, we select an equal amount of both classes.

{\scriptsize
\begin{lstlisting}[captionpos=b, caption=Axiom extraction, label=lst:axiomextraction, basicstyle=\ttfamily]
    SELECT ?class1 ?class2 ?label WHERE {
    { ?class1 a owl:Class . ?class2 a owl:Class . ?class1 rdfs:subClassOf ?class2
    filter (!isBlank(?class1)  && !isBlank(?class2))
    filter (?class1 != ?class2)
    bind(1.0 as ?label) }
    Union{ ?class1 a owl:Class . ?class2 a owl:Class . ?class1 owl:disjointWith ?class2
    filter (!isBlank(?class1)  && !isBlank(?class2))
    filter (?class1 != ?class2)
    bind(0.0 as ?label) }}
\end{lstlisting}}

The second step is to score, this is done by simply inputting the extracted axioms as a text file using the possibilistic heuristic~\cite{Tettamanzi2017}, and receiving an output file containing the scores (ARI), it should be noted that the process is very slow thus the need for a method such as ours. 

For disjointWith axioms, possibility only is considered since necessity is meaningless in the case of this axiom and incalculable. So the score is between 0 and 1.
\subsection{Semantic Measure Construction and Assignment}
\label{axiomsimilarity}
To be able to assign similarity measures between axioms, we need to retrieve the ontological distances between all classes. Using~\emph{Corese} in which the ontological distance metric is implemented, this translates into a function added to the~\emph{SPARQL} query. Query~\ref{lst:simextraction} retrieves three columns, the first two contain the combination of all classes with the third containing the ontological distance denoted by similarity. Blank nodes are ignored.

{\scriptsize
\begin{lstlisting}[captionpos=b, caption=Class ontological distance retrieval, label=lst:simextraction, basicstyle=\ttfamily]
    select * (kg:similarity(?class1, ?class2) as ?similarity) where {
    ?class1 a owl:Class . ?class2 a owl:Class
    filter (!isBlank(?class1)  && !isBlank(?class2)) }
\end{lstlisting}}

After retrieving the table of similarities it is pivoted to construct a symmetric~$n \times n$ matrix where the first column and the first row are the classes and the cells are the similarities between them with a diagonal of only~\textit{1's} since as we mentioned the similarity between a class~\textit{C} and itself is~\textit{1}. The shape of the concept similarity matrix is illustrated in Matrix~\ref{mtrx:conceptsim}.

\begin{figure}
{\scriptsize
\begin{equation*}
\begin{array}{c c c | c c c c c}
    & Concepts     &       &C_{0}        &C_{1}         & \dots    &C_{n} \\
    \hline   
    &C_{0} &        & 1           &  S_{0,1}     &  \dots   &S_{0,n}\\
    &C_{1} &        &S_{1,0}      & 1            &  \dots   &S_{1,n}\\
    &\vdots&        &\vdots       &\vdots        & \ddots   & \vdots\\
    &C_{n-1} &      & S_{n-1,0}   &S_{n-1,1}     & \dots    & S_{n-1,n}  \\
    &C_{n} &        & S_{n,0}     &S_{n,1}       & \hdots   & 1
\end{array}
\end{equation*}}
\caption{Concept similarity matrix}

\label{mtrx:conceptsim}
\end{figure}

\begin{figure}
{\scriptsize
\begin{equation*}
\begin{array}{c | c c | c c c c c}
Scores    & Axioms     &       &A_{0}        &A_{1}         & \dots    &A_{m} \\
    \hline   
score_{0}  &A_{0} &        & 1           &  S_{0,1}     &  \dots   &S_{0,m}\\
score_{1}  &A_{1} &        &S_{1,0}      & 1            &  \dots   &S_{1,m}\\
\vdots    &\vdots&        &\vdots       &\vdots        & \ddots   & \vdots\\
score_{m-1}  &A_{m-1} &      & S_{m-1,0}   &S_{m-1,1}     & \dots    & S_{m-1,m}  \\
score_{m}  &A_{m} &        & S_{m,0}     &S_{m,1}       & \hdots   & 1

\end{array}
\end{equation*}}
\caption{Axiom similarity matrix with labels}

\label{mtrx:axiomsim}
\end{figure}


From this matrix, we can derive the similarity between axioms. The goal is to end up with a similar square symmetric matrix of the shape $m \times m$, $m$ being the number of axioms, that has axioms instead of concepts as both first row and column, and the cells would be the similarities between a pair of axioms.
Exactly as in the case of the concept similarity matrix, the diagonal of this matrix will contain only~\textit{1's} as the similarity between an axiom~\textit{A} and itself is~\textit{1}.

\begin{algorithm}
\caption{Constructing the matrix of axiom similarities }\label{alg:Axiommatrixconstruction}
{\scriptsize
\begin{algorithmic}
\Require {\scriptsize All concepts included in axioms in $T_{A}$ be present in concept similarity matrix $M_{C}$}
\Ensure $0 \leq S \leq 1$\Comment{S is the similarity between 2 axioms}

\State $M_{C} \gets Concept\ similarity\ matrix$
\State $T_{A} \gets Set\ of\ labeled\ axioms$
\State $M_{A}\gets Axiom\ similarity\ matrix\ to\ be\ filled$

\For{$i = 0  \to{Length(T_{A}) - 1}$}
\For{$j = i  \to{Length(T_{A}) - 1}$}
\State $S_{L} \gets M_{C}[A_{i}[L],A_{j}[L]]$ 
\State $S_{R} \gets M_{C}[A_{i}[R],A_{j}[R]]$ 
\State $S^{1} \gets Avg(S_{L}, S_{R})$ \Comment{Experiments were done with (min, Avg)}
\If {Axiom Type is Disjointness or Equivalence}
\State $S_{L} \gets M_{C}[A_{i}[L],A_{j}[R]]$ 
\State $S_{R} \gets M_{C}[A_{i}[R],A_{j}[L]]$
\State $S^{2} \gets Avg(S_{L}, S_{R})$ 
\Else {}
\State $S^{2} \gets 0$ 
\EndIf
\State $S \gets Max(S^{1}, S^{2})$ 
\State $M_{A}.Add(A_{i}, A_{j}, S)$
\EndFor
\EndFor

\State\textbf{Note:} For Symmetric axioms, we have to compare both ways which is what we do inside the if statement for Disjointness and Equivalence axioms.

\end{algorithmic}
}

\end{algorithm}
In order to construct this axiom similarity matrix $M_{A}$ depicted in Matrix~\ref{mtrx:axiomsim} alongside the labels, we use Algorithm~\ref{alg:Axiommatrixconstruction}. The algorithm iterates over the set of labeled axioms $T_{A}$ which we extracted in Section~\ref{axiomextraction}, comparing each axiom $A_{i}$ to all other axioms $A_{j}$ in $T_{A}$ having $j$ increment from $i$ to length of $T_{A} - 1$ after each iteration of $i$. This is so we avoid redundant calculations. While comparing axioms $A_{i}$ and $A_{j}$, we first deal with the concept on the left side of each axiom, so the left concept of $A_{i}$ denoted by $A_{i}[L]$ and that of $A_{j}$ denoted by $A_{j}[L]$. To retrieve $S_{L}$ the similarity between those concepts from $M_{C}$, we search in the first row of the concept similarity matrix $M_{C}$ for concept $A_{i}[L]$, and in the first column of $M_{C}$ for concept $A_{j}[L]$. $M_{C}[A_{i}[L],A_{j}[L]]$, the intersection between the row where $A_{i}[L]$ was found, and the column where $A_{j}[L]$ was found represents the left side similarity between axioms $A_{i}$ and $A_{j}$. The same process is repeated for the right side, and then the axiom similarity $S$ is calculated as a function between $S_{L}$ and $S_{R}$. In this work we applied two functions and they were~\textit{minimum} and~\textit{average}. After calculating $S$ in each iteration of $j$ it is added to $M_{A}$ in the cell where the row is the index of $A_{i}$ and column the index of $A_{j}$.

\subsection{Axiom Base Vector Space Modeling}
\label{vectorspace}

We define a vector-space model to represent axioms as vectors. The number of dimensions $d$ of our vector space is equal to the number of axioms we have in the labeled set $T_{A}$. Each axiom can be represented as a vector $V$ in this $d$-dimensional space. Now that we have the similarities between axioms in our ontology, looking at the axiom-similarity matrix it would be intuitive to consider each row of the matrix as a vector $V$ in a vector space where the dimensions $d$ are the columns which are the axioms themselves having as weights the similarities $S$. Thus Algorithm~\ref{alg:Axiommatrixconstruction} is utilized whenever an axiom is generated or a new candidate axiom is suggested and will encode said axiom into a vector $V$ in this vector space.

\subsection{Score Prediction}
\label{regression}

After constructing our vector space and representing axioms as vectors, we consider the set of scored axioms as a set of vectors. It is now possible to apply regression machine learning methods on the vector space representation of an axiom base. We used a range of methods throughout our experiments such as random forests, Support vector regressor, and neural networks. Trees and random forests were mostly used to check that there was no bias during the prediction, since they allow us to interpret our predictive model easily and visually analyse the decisions.

To avoid information leakage since the matrix is symmetric, considering the size of the matrix is  $n \times n$, all models would be trained using an $m \times m$ sub-matrix of the axiom similarity matrix with the labels, and then tested on a $z \times m$ sub-matrix. This means an axiom vector $V$ will have $m$ dimensions (features) which are the ones used to train the model, regardless what the number of dimensions (features) in the full matrix is. In other words the symmetric part of the matrix equivalent to $z$ (the columns of $n$ outside the range of $m$) is discarded. The model's goal is to predict the score of a candidate axiom, which is represented as a vector $V$ in our vector space having $m$ dimensions (features), which are the axiom's similarities with the axioms of the same type used to train the model. The score is a number between -1 and 1 for subClassOf and equivalence, and between 0 and 1 for disjointWith. where $1$ represents the highest acceptability. Any scorer/scoring algorithm can be used, we decided on~\cite{Tettamanzi2017} to be able to compare our results with~\cite{Malchiodi2018}.

\section{Experiments and Results}
\label{experiments}

For the experiments\footnote{All the data and code needed to replicate the experiments available at https://anonymous.4open.science/r/axiom-score-prediction-BC16}, the workstation that was used had the following hardware configuration:
{\small
  \begin{itemize}
    \item Dual CPUs: 2 $\times$ Intel(R) Xeon(R) CPU E5-2689 0 @ 2.60GHz base and 3.30 all core boost. With 8 cores and 16 threads per CPU for a total of 16 cores and 32 threads.
    \item A total of 32 GB of RAM memory with frequency 1600 MHZ distributed as 8 $\times$ 4 GB sticks with 4 sticks assigned to each CPU.
    \item 1 TB of NVME SSD storage with read and write speeds of up to 2000 MB per second.
  \end{itemize}
  }
\subsection{Dataset Preparation}
\label{dataprep}

For our testing and experiments, and to comply with both the scorer and the experiment of~\cite{Malchiodi2018}, the ontology used is Dbpedia. This also allows us to show how our method would perform in a real-world case.

 We used two datasets for our experiments, one for axiom type subClassOf and it is the one used in~\cite{Malchiodi2018}, and another for axiom type disjointWith which is a generated set of atomic disjointWith candidate axioms, as described in Sect.~\ref{axiomextraction}, scored using~\cite{Tettamanzi2017}.

For the axiom set used in~\cite{Malchiodi2018}, they have 722 subClassOf axioms and their negations, making it 1444 axioms. The negations are specific for that similarity and for the support vector clustering method that was applied in~\cite{Malchiodi2018}. These negations serve no meaning if it is possible to predict the scores of the original 722 axioms. For this reason, and since we had the possibility of an axiom and its negation, we can use that to calculate the ARI using~\ref{eq:ARI}. Then the dataset we work this will be the 722 axioms with their ARI score, since we will not be using the same machine learning method nor similarity for our proposed solution.

We did not include tests of equivalentClass axiom types since the process of creating the axiom similarity matrix of that axiom type is the same as disjointWith. They are both symmetrical axioms and equivalentClass axioms are basically a two way subClassOf axiom. For that reason we decided to include our experiment on the type disjointWith.
  
 {\small
  \begin{table*}[ht]

    \centering
    \begin{supertabular}{c c c | c c | c c c c c c c c c}
      Similarity Method &Type& Set size & ASM & Candidate processing& NN & Random Forest & SVR & Modified SVC & Explained variance \\
     \hline
     Instance based&subClassOf& 1444 &   649,440     &  1,799&  0.35299    & 0.30707  & \textbf{0.26721} & 0.572& 0.52652 \\
     Ontological based&subClassOf&722    & \textbf{13.7275} &    \textbf{0.01901}  &   \textbf{0.31442}   & \textbf{0.30231} & 0.33972& NA& \textbf{0.88859}  \\
    
     Ontological based&disjointWith&3868 &   129.9637   &   0.03359   & 0.23325 & 0.21754  & 0.23771 & NA & 0.73462   \\
    \end{supertabular}
    
    \caption{\scriptsize Time cost in seconds, performance scores in RMSE per model, and the explained variance score of the best performing model per experiment.}
    \label{tab:steptimes}
    
  \end{table*}}
  
  For the set of axioms of type disjointWith, we load the Dbpedia OWL file into Corese. The OWL file contains no individual data which means faster processing in the search engine, corese reasoner was applied to it to deduce the maximum number of axioms. Positively scored disjointness axioms were obtained from the results of~\cite{Nguyen2019}.\footnote{ https://bitbucket.org/RDFMiner/classdisjointnessaxioms/src/\\master/Results/ClassDisjointnessAxioms/} Negatively scored disjointWith axioms were existing subClassOf axioms, explained in Section~\ref{axiomextraction}. In case the used ontology is already populated with explicit axioms, this would be done in an attempt to obtain a set of axioms that the scorer will not provide a score close to 0 i.e., a score that implies ignorance for lack of instance data. This allows us to learn a better model that makes better predictions and surpasses the limitations a statistical scorer can face. But in our case this would result in a small set of 1120 scored axioms. One of the issues the authors of~\cite{Malchiodi2018} faced is that the dataset they were working with was too small and they point out that this had prevented them from running certain experiments. For this case we combined both approaches expressed in Section~\ref{axiomextraction}. Randomly generating atomic candidates and checking that they do not already exist, we managed to score an additional 2748 axioms for a total of 3868. This will allow better testing to see how the method performs in real-world cases and on a different axiom type.
  
  We will denote by $T_{A}$ the axiom set being used. We will not be comparing the time needed to score the set of axioms since they both use the same scorer. Heuristic~\cite{Tettamanzi2017} takes a list of candidate axioms as input and provides as output a list of scored candidate axioms. It is a slow but precise heuristic that depends on instance data and counting queries.
  
After preparing the set of axioms, we have to produce the concept similarity Matrix~\textbf{(CSM)}~\ref{mtrx:conceptsim}, a process we detailed in Section~\ref{similarity}. The processing time for creating the~\textbf{(CSM)} is dependent on the number of concepts. In our tested dataset, the number of unique concepts was 762 and creating its~\textbf{(CSM)} took 2.0362 seconds.

 The next step is constructing the axiom similarity matrix \textbf{(ASM)}, and encoding each of the axioms as vectors $V$ in our vector space. This step is detailed in the second half of Section~\ref{axiomsimilarity} and in Section~\ref{vectorspace}. The algorithm applied to our axiom data set $T_{A}$ is Algorithm~\ref{alg:Axiommatrixconstruction}, this is the same algorithm used to encode candidate axioms into the vector space, it is the equivalent of running queries~\ref{eq:querysim} and~\ref{eq:querysim2} for the method proposed in~\cite{Malchiodi2018} to retrieve the similarity. In contrast the algorithm we use is not based on instance counting and is much faster in comparison. Table~\ref{tab:steptimes} details the time required to complete the axiom similarity matrix. As observed the time needed to complete the task of building the \emph{ASM} significantly increases as the number of axioms processed increases, but we also know that it depends on the size of the \emph{CSM} as well from the time needed to encode a single axiom into a vector. As the number of concepts increases the \emph{CSM} being searched increases as it has the shape $n_{concepts} \times n_{concepts}$ and the number of cells $n^{2}$. We would note that the test scenario (3868 disjointWith axioms) presented in Table~\ref{tab:steptimes} is extreme, especially the case of Dbpedia, as the number of axioms needed for training a model does not need to be as large to achieve peak accuracy/results. It is possible with a dataset size of subClassOf experiment (722).
  
  While constructing the ASM using the instance similarity method, we had to calculate the denominators of all axiom pairs as in Equation~\ref{eq:denom}. We ran into an issue of lack of instance data, this means a 0 in the denominator which is obviously a huge problem. The authors of~\cite{Malchiodi2018} address this by denoting any similarity between a pair of axiom where the denominator is 0 (lack of instances) with a 0 as well. This highlights the weakness of instance based similarity methods. We do not believe that simply saying the similarity between a pair of axioms is 0 because they do not share individual data.

\subsection{Training and Testing}
\label{testing}

After obtaining our datasets as described in the previous section, we applied different regression methods as mentioned in Section~\ref{regression} to check how the similarity measure performs. 

\setcounter{figure}{0}  
\renewcommand\figurename{Figure}


The methods we tested include, but are not limited to, Random Forests, Neural Networks and Support Vector Regression. Experiments were with both Average and Minimum functions to get the similarity $S$ between axioms, from $S_{L}$ and $S_{R}$ (as explained in Section~\ref{axiomsimilarity}), we will denote by ASF the axiom similarity function here on out. All this means that the number of experiments performed was large so we will only report on some scenarios.

We used RMSE (root mean squared error) as the score, to be consistent with~\cite{Malchiodi2018} during the comparison. We applied hyper-parameter tuning using grid search as well as five fold cross validation to infer better models. Table~\ref{tab:steptimes} shows the best results (in terms of RMSE) for each experiment using each model.

While replicating the experiment of~\cite{Malchiodi2018}, we decided to test other methods in addition to theirs. We will use the authors' best result for the comparison with the original model. We will be using the original 722 for our proposed method, since our similarity does not require negated axioms (no need for the extra 722 negated axioms).

\subsection{Results and Observations}
\label{results}

Comparing ASM creation and candidate axiom encoding time costs in Table~\ref{tab:steptimes}, for both our proposed method and the one in~\cite{Malchiodi2018}, our ontological based similarity seems almost instantaneous. For the instance based similarity it is a problem during dataset preparation as it needs seven and a half days to prepare a data set of 722 axioms (and their negations), as well as candidate axiom encoding/processing where it takes half an hour to process every candidate axiom we want to predict a score to, whereas our ontological based method requires about fourteen seconds to process and prepare the exact same dataset, and less than 0.1 s to encode a new axiom, making it much more preferable with regards to time cost.

During the experiments we were unable to compare both methods in anything other than subClassOf axioms. This is because the method detailed in~\cite{Malchiodi2018} can only handle subClassOf axioms making it very limited and constrained, whereas our proposed method can address all atomic class axiom types, all that is needed is training one model for each set. Considering the timing needed to prepare the training data (ASM), as seen in Table~\ref{tab:steptimes}, this is easily doable and very efficient.

We observe in Table~\ref{tab:steptimes} that the time cost for creating the disjointWith (129 seconds for 3868 axioms) experiment's ASM was almost ten times that of the subClassOf (13 seconds for 722 axioms), even though the number of axioms is not ten times as much, only about five. This is normal since disjointWith axioms are symmetrical, and as shown in Algorithm~\ref{alg:Axiommatrixconstruction} and explained in Section~\ref{axiomsimilarity}, the calculation is doubled for every axiom to check forwards and backwards the similarity between the pair of axioms.

It is very important to note that the instance-based similarity method performed better when used with a machine learning method different from the modified support vector clustering method used in~\cite{Malchiodi2018}. It was able to achieve less than half the RMSE score of what was stated in~\cite{Malchiodi2018}. This suggests that the modified support vector clustering method is unnecessarily complicated and not the most appropriate method to deal with this problem, which turns out to be relatively easy once a good similarity measure manages to transform it to a suitable representation. The ontological based method outperformed the instance based method in Neural Network and Random forests models. For the same data set, the subClassOf set, the two methods seem to perform closely in terms of RMSE, while the ontological distance method seems to be achieving very good scores in the larger disjointWith set. This made us look more in-depth to what was happening. Turning our attention to the support vector regression model, we can see in Table~\ref{tab:steptimes} that for the same dataset (subClassOf) the instance based measure performed considerably better than the ontological based method, but was still outperformed in the larger (disjointWith) set as far as performance goes. So we investigated the reason behind this specific performance for this specific dataset in this model, and found the best explanation in the explained variance score. We found that for the instance based method, according to the explained variance score of~\textbf{0.56252} for its best performing model the support vector regressor, this performance is specific to this dataset. This means there exists a bias in the training set and it is expected that any new candidate will suffer from a higher error rate than what is experienced in the training stage. This is explained by the fact that both similarity method and dataset were handcrafted and picked to work with a support vector clustering method modified to work as a regressor. On the other hand for the ontological based method the explained variance score throughout all our experiments ranged between 0.71 up to 0.88 meaning this method will produce better results when predicting scores for new candidate axioms.

All results shown in Table~\ref{tab:steptimes} were obtained using the Average ASF, as it outperformed the Minimum ASF in all our runs.

\section{Conclusion}

We have proposed a method with the aim of learning predictors for the acceptability of an atomic candidate \emph{OWL} axiom of any type. The method relies on a semantic similarity measure derived from the ontological distance between concepts in a subsumption hierarchy. Extensive tests that covered multiple parameters and settings were carried out to investigate the effectiveness and potential of the method compared with the state of the art. 

The results obtained strongly support the effectiveness of the proposed method in predicting the scores of the considered OWL axiom types with a consistently low error rate. This allows us to confidently say that our proposed method can be used in combination with ILP or statistical methods, such as Dl-learner~\cite{BUHMANN201615} or a Grammatical evolution approach such as~\cite{Nguyen2019a}, as a building block or extension/plugin to allow faster execution while maintaining accuracy.

We attribute the high accuracy and extremely good performance of our method to the close relation between the similarity measure and the model-theoretic semantics of axioms. This merits further investigation of links between similarity/distance measures and semantics, and the effect on performance of machine learning models in logical reasoning tasks.

Based on our findings, some research paths emerge, including:
{\small
\begin{itemize}

\item Developing the method to predict the scores of complex candidate axioms.
\item Extending the method to property axioms since they also possess an~\textbf{IS-A} hierarchy, which is the base of our similarity measure.
\item Considering an ensemble approach that incorporates active learning based of the agreement or disagreement of included models based on a threshold such as the variance.

\end{itemize}
}




\end{document}